\documentclass[lettersize,journal]{IEEEtran}
\usepackage{amsmath,amsfonts}
\usepackage{algorithmic}
\usepackage{algorithm}
\usepackage{array}
\usepackage[caption=false,font=normalsize,labelfont=sf,textfont=sf]{subfig}
\usepackage{textcomp}
\usepackage{stfloats}
\usepackage{url}
\usepackage{verbatim}
\usepackage{graphicx}
\usepackage{cite}

\usepackage{adjustbox}
\usepackage{colortbl}  
\usepackage{xcolor}
\usepackage{array}
\usepackage{booktabs}
\usepackage{times}
\usepackage{soul}
\usepackage{url, listings,amssymb}
\usepackage[hidelinks]{hyperref}
\usepackage[utf8]{inputenc}

\usepackage{graphicx}
\usepackage{amsmath}
\usepackage{amsthm}
\usepackage{booktabs}
\usepackage{algorithm}
\usepackage{algorithmic}
\usepackage[switch]{lineno}
\usepackage{multirow}

\begin{document}

\title{EAVL: Explicitly Align Vision and Language for Referring Image Segmentation}

\author{Yichen Yan, Xingjian He, Wenxuan Wang, Sihan Chen, Longteng Guo, Jing Liu\textsuperscript{\dag},~\IEEEmembership{Member,~IEEE,}
\thanks{This work was supported by the National Science and Technology Major Project (No.2022ZD0118801), National Natural Science Foundation of China (U21B2043, 62206279)}
\thanks{Y. Yan, X. He, W. Wang, Si. Chen, L. Guo and J. Liu are with the Institute of Automation, Chinese Academy of Sciences, Beijing 100190, China, while Y. Yan, W. Wang, S. Chen and J. Liu are also with the School of Artificial Intelligence, University of Chinese Academy of Science, Beijing 100190, China. (e-mail: yanyichen21@mails.ucas.ac.cn, xingjian.he@nlpr.ia.ac.cn, wangwenxuan2023@ia.ac.cn, sihan.chen@nlpr.ia.ac.cn, jliu@nlpr.ia.ac.cn, longteng.guo@nlpr.ia.ac.cn)}

\thanks{\textsuperscript{\dag}Corresponding author.}}



\maketitle

\begin{abstract}
Referring image segmentation (RIS) aims to segment an object mentioned in natural language from an image. The main challenge is text-to-pixel fine-grained correlation. In the previous methods, the final results are obtained by convolutions with a fixed kernel, which follows a similar pattern as traditional image segmentation.  These methods lack explicit alignment of language and vision features in the segmentation stage, resulting in suboptimal correlation. In this paper, we introduce EAVL, a method explicitly aligning vision and language features. In contrast to fixed convolution kernels, we introduce a Vision-Language Aligner that aligns features in the segmentation stage using dynamic convolution kernels based on the input image and sentence. Specifically, we generate multiple queries representing different emphases of language expression. These queries are transformed into a series of query-based convolution kernels, which are applied in the segmentation stage to produce a series of masks. The final result is obtained by aggregating all masks. Our method harnesses the potential of the multi-modal features in the segmentation stage and aligns language features of different emphases with image features to achieve fine-grained text-to-pixel correlation.  We surpass previous state-of-the-art methods on RefCOCO, RefCOCO+, and G-Ref by large margins. Additionally, our method is designed to be a generic plug-and-play module for cross-modality alignment in RIS task, making it easy to integrate with other RIS models for substantial performance improvements.
\end{abstract}

\begin{figure}[t]
  \centering
  \includegraphics[width=\linewidth]{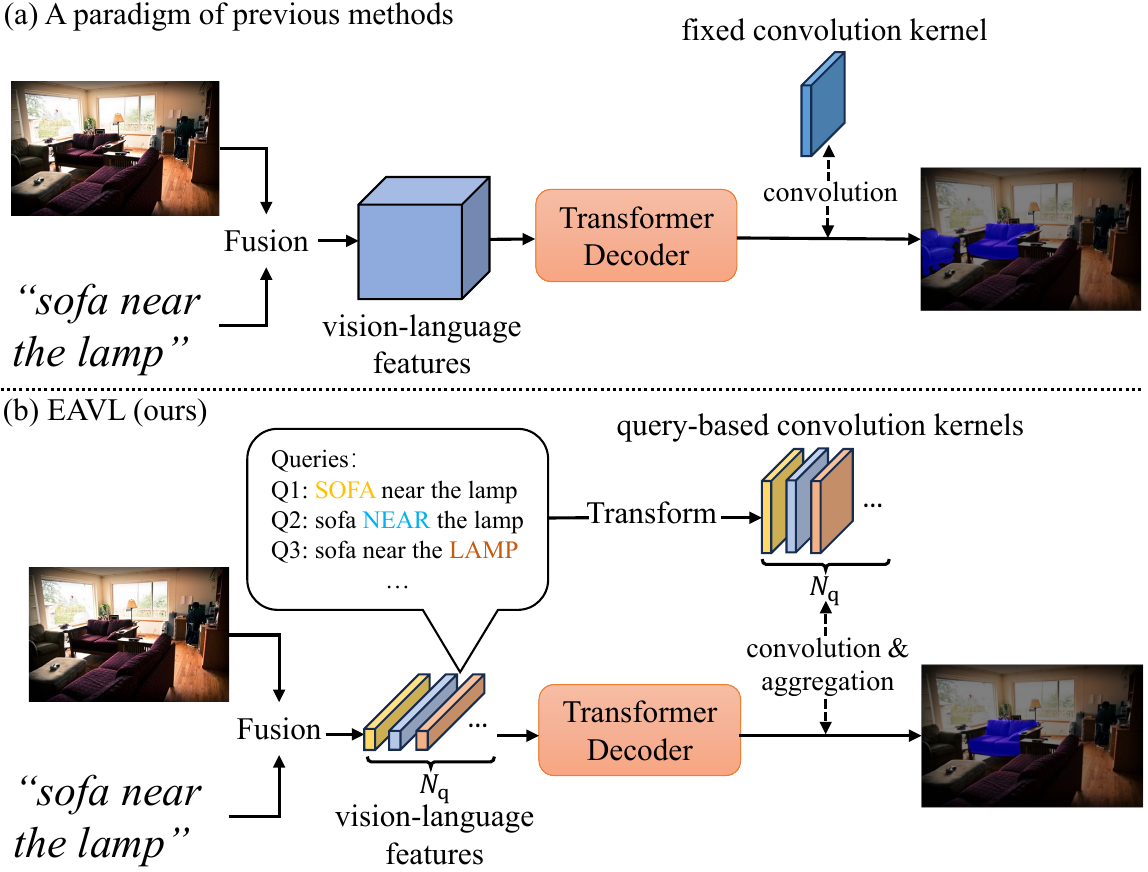}
  \caption{(a) In the previous methods (i.e., VLT~\cite{ding2021vision}), 
 the vision-language features obtained after fusion are directly fed into a Transformer Decoder, and the final result is obtained using a fixed convolution kernel. This approach is similar to the method employed in the segmentation stage of traditional image segmentation. (b) Our approach differs from previous methods by generating specialized vision-language features known as queries and transforming them into a series of dynamic query-based convolution kernels. Our method not only maximizes the potential of vision-language features but also explicitly aligns the vision features with language features to achieve text-to-pixel fine-grained correlation.}
 \label{fig:motivation}
\end{figure}


\section{Introduction}
\IEEEPARstart{R}{eferring} image segmentation (RIS)~\cite{hu2016segmentation,liu2017recurrent}, which aims to segment an object referred by natural language expression from an image, is a fundamental vision-language task.  Unlike traditional segmentation tasks such as semantic segmentation~\cite{pan2024purify,zhuang2020video,ji2020encoder,tian2021partial}, and instance segmentation~\cite{yang2021task,zhang2021segmenting,yan2022solve,wang2023efficient} which only handle predefined categories, RIS offers greater flexibility by allowing segmentation of target objects from any category and location as specified in the language input.  So this task has numerous potential applications, including interactive image editing and human-object interaction ~\cite{wang2019reinforced}. Referring image segmentation poses a major challenge called text-to-pixel fine-grained correlation. Text-to-pixel fine-grained correlation is to correlate vision and relevant language information at the pixel level.

Early works have primarily focused on how to fuse vision and language features effectively, with a common approach that utilizes concatenation and convolution to fuse them. With the advent of attention mechanism, several methods~\cite{ding2021vision,feng2021encoder}  adopt the vision-language attention mechanism to learn cross-modal features more effectively. Recently, some methods~\cite{wang2022cris,yang2022lavt} have started using pre-trained models to align and fuse the features. Prior methods have primarily emphasized the enhancement of cross-modal feature fusion. They use convolutions with a fixed kernel in the segmentation stage which implicitly aligns the vision and language features. However, addressing text-to-pixel fine-grained correlation requires explicit alignment of vision and language features. This explicit alignment ensures that the image and language are mutually influenced and contribute to accurate results.

As illustrated in Fig.~\ref{fig:motivation}(a), previous methods extract visual features from images and textual features from language expressions separately. These features are then fused into vision-language features. However, in the subsequent stage, these methods follow a conventional approach by directly inputting the fused features into a Transformer~\cite{vaswani2017attention} decoder and applying convolution operations with a fixed convolution kernel. This traditional approach has two drawbacks: (1). It only focuses on the fusion process of obtaining the vision-language features without fully harnessing their inherent potential in the segmentation stage. (2). Regardless of the content of language and image input, a fixed convolution kernel is utilized to generate the final prediction mask, resulting in only implicit alignment of vision and language features in the segmentation stage.

In this paper, we argue that the fused vision-language features have untapped potential to guide explicit alignment in the segmentation stage. In  Fig.~\ref{fig:motivation}(b), we introduce a method that uses a series of dynamic convolution kernels generated based on the input language and image.  To accomplish this, we  extract vision and language features respectively, and fuse them into a series of queries. Each query represents a different emphasis of the input sentence. These queries, also denoted as vision-language features, are not only fed into a Transformer Decoder as usual but also transformed into a series of dynamic query-based convolution kernels. These multiple dynamic kernels are then used to perform convolutions with the output of the transformer Decoder, resulting in a series of corresponding segmentation masks. The final result is obtained by the aggregation of these masks. This method of generating different convolution kernels depending on the inputs enables explicit alignment of vision and language features in the segmentation stage. Our method maximizes the potential of multi-modal features and explicitly aligns the vision features with language features of different emphasis. Moreover, this explicit alignment facilitates text-to-pixel fine-grained correlation and obtains accurate results by aggregating multiple masks obtained through query-based convolution kernels.

\begin{figure*}[t]
  \centering
  \includegraphics[width=\linewidth]{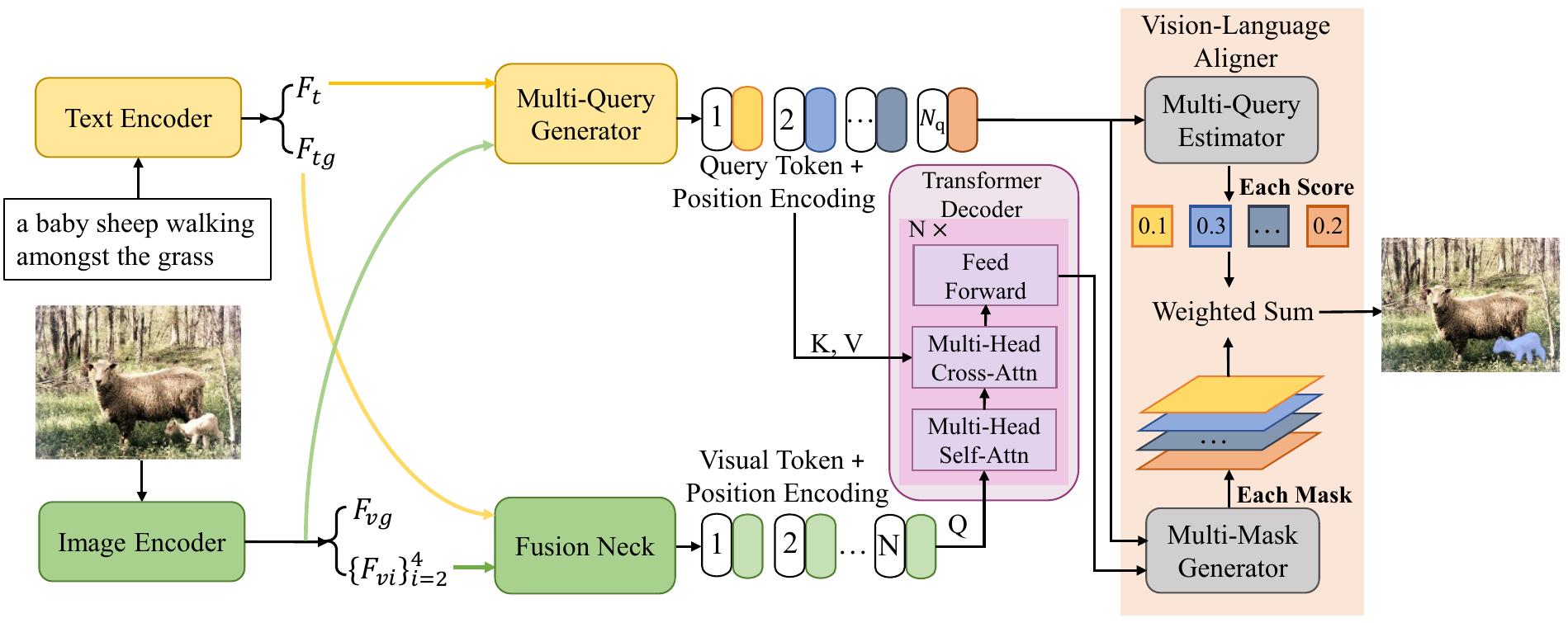}
  \caption{EAVL mainly consists of a
text encoder, an image encoder, a Multi-Query Generator, a Transformer Decoder, and a Vision-Language Aligner. The Vision-Language Aligner has two parts, a Multi-Mask Generator and a Multi-Query Estimator.}
\label{fig:method}
\end{figure*}

 As illustrated in Fig.~\ref{fig:method}, We use vision and text encoder to extract image features and language features, then we generated multiple queries and each query represents a different emphasis of the input sentence. These queries are not only fed into a Transformer Decoder like the previous approach but also sent to a Vision-Language Aligner which includes a Multi-Mask Generator and a Multi-Query Estimator. In Multi-Mask Generator, these queries are transformed into a series of query-based convolution kernels to do convolution with the outputs of the Transformer Decoder. Therefore, each query will produce a mask representing a different emphasis of the language expression. These queries will be also used to generate a series of scores that represent their importance in the Multi-Query Estimator. The final result is obtained by computing the weighted sum of all the masks. Through this approach, we explicitly align language features of different emphases with the image features. Overall, our novel structure enhances the exploitation of these queries, and these dynamic query-based convolution kernels achieve an explicit alignment of the vision and language features, resulting in an effective text-to-pixel fine-grained correlation and accurate results. Moreover, our Vision-Language Aligner design can be applied in other transformer-based RIS methods and obtain significant improvements, indicating that our design is a generic approach in cross-modality alignment. 
 In summary, our main contributions are listed as follows:

\begin{itemize}
    \item We introduce EAVL, a novel framework for referring image segmentation. Our approach employs a Vision-Language Aligner in the segmentation stage, explicitly aligning vision and language features to effectively address text-to-pixel fine-grained correlation.
    \item We generate a series of queries, which are also vision-language features. Instead of only sending them into a Transformer Decoder as usual, we transform them into query-based convolution kernels and produce corresponding masks to obtain the final results. This approach allows us to not only acquire vision-language features but also effectively exploit their potential to enhance the overall performance of the model.
    \item  The experimental results on three challenging benchmarks significantly outperform previous state-of-the-art methods by large margins. Further experiments show that our proposed Vision-Language Aligner can be applied to other existing transformer-based RIS models, leading to significant improvements
\end{itemize}

\section{Related Works}

 Referring image segmentation aims at segmenting a specific region in an image by comprehending a given natural language expression~\cite{hu2016segmentation}. Given a natural language expression describing the properties of the target object in the given image, the goal of the referring segmentation is to ground the target object referred by the language and generate a corresponding segmentation mask.  With the wide application of deep learning techniques in vision-language tasks~\cite{zhang2020language,yu2019multimodal,miao2024temporally,liu2020improving},  RIS has become
one of the most fundamental and challenging cross-modal tasks.  It has the potential to be applied in a wide range of domains, such as interactive image editing and human-object interaction. This task is first introduced by Hu et al~\cite{hu2016segmentation}.

Early works~\cite{hu2016segmentation, li2018referring,liu2017recurrent} initially extracted visual features using Convolutional Neural Networks (CNNs) and language features using Long Short-Term Memory networks (LSTMs). These vision and language features were then fused through concatenation, and the resulting fused features were processed by a fully convolutional network (FCN)\cite{long2015fully} to generate the target segmentation mask. Liu et al.\cite{liu2017recurrent} advanced this by concatenating visual features with the features of each word, creating a multi-modal feature sequence that was recurrently refined using ConvLSTM. Li et al.\cite{li2018referring} further improved performance by fusing high-level multi-modal features with lower-level, higher-resolution visual features multiple times. Margffoy-Tuay et al. introduced DMN\cite{margffoy2018dynamic}, employing dynamic filters to enhance cross-modal interaction. Chen et al.~\cite{chen2019referring} took a different approach by employing a caption generation network to produce a sentence describing the target object and enforcing consistency between the generated caption and the input expression. Additionally, MAttNet~\cite{yu2018mattnet} proposed a two-stage method, leveraging Mask-RCNN~\cite{he2017mask} to detect all possible objects in an image, followed by selecting the object most relevant to the text. CMPC~\cite{huang2020referring} constructed a fully connected graph over the multi-modal features and applied graph convolution to reason about the final location.

Although simple and effective, early frameworks based on CNN and LSTM models have notable limitations in handling complex language expressions. When encoding language descriptions sequentially, these methods overlook the varying dependencies and importance of different sentence components. This can lead to a biased understanding, as irrelevant or less important words may introduce noise and confuse the model. Furthermore, due to the local receptive fields of CNNs and the sequential processing nature of LSTMs, these algorithms often struggle to capture global contextual information, leading to an incomplete understanding of the overall target. This limitation becomes evident in their inability to fully grasp the surrounding environment of the target, which may result in unclear object boundaries or missed key details in segmentation results.

As the attention mechanism and transformer arouse more and more interest, a series of works are proposed to adopt the attention mechanism to fuse the vision and language features. For example, In CMSA~\cite{ye2019cross}, a cross-modality self-attention module is employed to build long-range dependency between two modalities. VLT~\cite{ding2021vision} employs a transformer to build a network with an encoder-decoder attention mechanism for enhancing the global context information. LAVT~\cite{yang2022lavt} conducts multi-modal fusion at intermediate levels of the transformer-based network. CRIS~\cite{wang2022cris} leverages pre-trained CLIP~\cite{wang2022clip} and transformer decoder to transfer the CLIP's knowledge from text-to-image matching to text-to-pixel matching. $\mathbf{M^3}$AII~\cite{liu2023multi} utilizes Multi-Modal Mutual Attention to effectively fuse information from the two input modalities. In order to avoid abundant vision features, FCNet~\cite{yan2024fuse} employs cross-modal attention to capture the vision emphasis features and feed these features into a transformer to facilitate key information propagation. CRFormer~\cite{Yan_2024} proposes a calibration transformer decoder that can iteratively calibrate the language features to help deep vision-language integration.
RefSegformer~\cite{wu2023towards} introduces a novel task where the described objects may not be present in the image. However, previous work focuses on how to improve cross-modal feature fusion effectively, without fully exploiting the potential of the fused vision-language features. Moreover, they use the fixed convolution kernel which only implicitly aligns the vision and language features, resulting in a suboptimal text-to-pixel fine-grained correlation. On the contrary, our proposed method effectively utilizes the fused features to generate dynamic convolutional kernels to facilitate explicit vision-language alignment.

\section{Methodology}
As illustrated in Fig.~\ref{fig:method}, our proposed framework facilitates knowledge transfer to generate multiple queries and their corresponding masks to obtain the final prediction. Firstly, the framework takes an image and a language expression as input. We leverage an image encoder and a text encoder to extract image and text features, respectively. These extracted features are then combined in the Fusion Neck to obtain the simple vision-language features. After that, we utilize the vision feature and the language features to generate multiple queries in the Multi-Query Generator. 

Then, the generated queries and simple vision-language features are input into the Transformer Decoder. The outputs of the decoder, along with the generated queries, are fed into the Multi-Mask Generator to produce multiple masks. Meanwhile, the Multi-Query Estimator uses the queries to determine the weights of each mask. Finally, we use those masks and their corresponding weights to calculate the weighted sum, obtaining the final result.  

\subsection{Image and Text Feature Extraction}

\textbf{Text Encoder}. For a given language expression $T \in \mathbb{R}^{L}$, we utilize a BERT~\cite{devlin2018bert} to obtain text features and transform it into $F_t \in \mathbb{R}^{L\times C}$ by a linear projection. Additionally, we use the highest layer's activations of the Transformer at the \texttt{[CLS]} token as the global feature for the entire language expression. This feature is linearly transformed and denoted as $F_{tg} \in \mathbb{R}^{1 \times C}$. Here, $C$ represents the feature dimension, while $L$ is the length of the language expression.

\textbf{Image Encoder}. For a given image $I \in \mathbb{R}^{H \times W \times 3}$, we extract not only local but also global visual representations. As an example, we use the Swin Transformer~\cite{liu2021swin} encoder. There are 4 stages in total and we denote the features as $\left\{\mathbf{x}_{i}\right\}_{i=1}^{4}\in \mathbb{R}^{H_{i} \times W_{i} \times C_i}$. Here, $H_i$, $W_i$, and $C_i$ are the height, width, and number of channels of $\mathbf{x}_i$. Unlike the original Swin Transformer, we add an attention-pooling layer. Specifically, we applies global average pooling to $\mathbf{x}_{4}$ to obtain a global feature, denoted as $\overline{\mathbf{x}}_4 \in \mathbb{R}^{C_4}$. Then, we concatenates the features $\left[\overline{\mathbf{x}}_4, \mathbf{x}_4\right]$ and feeds them into a multi-head self-attention layer.

\begin{equation}
[\overline{\mathbf{z}}, \mathbf{z}]=MHSA\left(\left[\overline{\mathbf{x}}_4, \mathbf{x}_4\right]\right).
\end{equation}
Here, we obtain output $\mathbf{z}$ as a feature map due to its adequate spatial information. We also get the global vision feature $\overline{\mathbf{z}}$. 

In our proposed model, we leverage the multi-scale vision features $\mathbf{x}_{2} \in \mathbb{R}^{H_{2} \times W_{2} \times C_2}$ and $\mathbf{x}_{3} \in \mathbb{R}^{H_{3} \times W_{3} \times C_3}$ from the second and third stages, respectively. We denote them as $F_{v 2} \in \mathbb{R}^{H_{2} \times W_{2} \times C_2}$ and $F_{v 3} \in \mathbb{R}^{H_{3} \times W_{3} \times C_3}$, respectively. In the fourth stage, different from the original ResNet, we use not only the vision features $\mathbf{z}$ for its sufficient spatial information but also incorporate the global vision feature $\overline{\mathbf{z}}$ to capture the global information of the image. We denote $\mathbf{z}$ and $\overline{\mathbf{z}}$ as $F_{v 4} \in \mathbb{R}^{H_{4} \times W_{4} \times C_4}$ and $F_{v g} \in \mathbb{R}^{C_4}$. Eventually, we extract $\left\{F_{vi}\right\}_{i=2}^{4}$ and $F_{v g}$ from the image.

\textbf{Fusion Neck}. In the Fusion Neck, we perform a simple fusion of vision and language features, including $F_{v 2}$, $F_{v 3}$, $F_{v 4}$, and the global language feature $F_{t g}$, to generate a vision feature that incorporates global language information. Initially, we fuse $F_{v 4}$ and $F_{t g}$ to obtain $F_{m 4} \in \mathbb{R}^{H_{3} \times W_{3} \times C}$ using the following equation:

\begin{equation}
F_{m 4}=U p\left(\sigma\left(F_{v 4} W_{v 4}\right) \cdot \sigma\left(F_{tg} W_{tg}\right)\right),
\label{eq:2}
\end{equation}
In this process, $Up(\cdot)$ denotes 2 $\times$ upsampling function, and $\cdot$ denotes element-wise multiplication. We first transform the visual and textual representations into the same feature dimension using two learnable matrices, $W_{v 4}$ and $W_{tg}$, and then We apply ReLU activation function which is denoted as $\sigma$ to generate $F_{m 4}$. Subsequently, we obtain the multi-modal features $F_{m 3}$ and $F_{m 2}$ using the following procedures:

\begin{equation}
\begin{aligned}
& F_{m_3}=\left[\sigma\left(F_{m_4} W_{m_4}\right), \sigma\left(F_{v_3} W_{v_3}\right)\right], \\
& F_{m_2}=\left[\sigma\left(F_{m_3} W_{m_3}\right), \sigma\left(F_{v_2}^{\prime} W_{v_2}\right)\right], \\
& F_{v_2}^{\prime}=Avg\left(F_{v_2}\right),
\end{aligned}
\end{equation}

Where $Avg(\cdot)$ denotes a $2\times2$ average pooling operation with 2 strides, [,] denotes the concatenation operation.Subsequently, we concatenate the three multi-modal features ($F_{m 4}, F_{m 3}, F_{m 2}$) and use a $1 \times 1$ convolution layer to aggregate them:
\begin{equation}
F_m= Conv\left(\left[F_{m_2}, F_{m_3}, F_{m_4}\right]\right),
\end{equation}

Where $F_{m} \in \mathbb{R}^{H_{3} \times W_{3} \times C}$. Then, we use the 2D spatial coordinate feature $F_{coord} \in \mathbb{R}^{H_{3} \times W_{3} \times 2}$ and concatenate it with $F_{m}$, flattening the result to obtain the fused vision features with global textual information, denoted as $F_{vt} \in \mathbb{R}^{H_{3}W_{3} \times C}$.
\begin{equation}
\begin{aligned}
& F_{inte}=Conv\left(\left[F_m, F_{coord}\right]\right),\\
& F_{vt}=Flatten\left(F_{inte}\right).
\end{aligned}
\end{equation}
Here, $Flatten(\cdot)$ denotes flatten operation and $F_{inte} \in \mathbb{R}^{H_{3} \times W_{3} \times C}$ is intermediate features. After the flatten we obtain the $F_{vt} \in \mathbb{R}^{N \times C}$, $N = H_{3} \times W_{3} = \frac{H}{16} \times \frac{W}{16}$, which will be utilized in the following process. 
\begin{figure}[t]
  \centering
  \includegraphics[width=\linewidth]{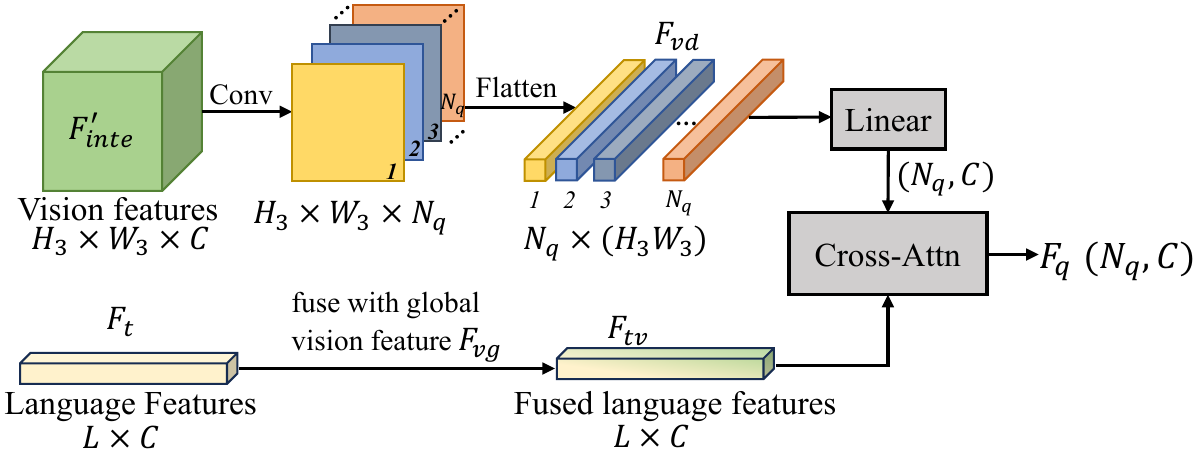}
 \caption{The details of Query Generation process. }
 \label{fig: query generation}
\end{figure}
\subsection{Multi-Query Generator} We use a cross-attention mechanism to generate new queries. Each query represents a different emphasis and will be transformed into a series of query-based convolution kernels in the Multi-Mask Generator. 

As shown in Fig.~\ref{fig:method}, the Multi-Query Generator takes multi-scale visual features $\left\{F_{vi}\right\}_{i=2}^4$, the global vision feature $F_{vg}$, and language features ${F_t}$ as input and outputs a series of queries. To generate multiple queries, we first need to obtain dense vision features and fused language features. The detailed process is shown in Fig.~\ref{fig: query generation}.
  
 \textbf{Dense Vision Features}. In order to obtain the dense vision features, we use the multi-scale vision features $\left\{F_{vi}\right\}_{i=2}^4$. The operations we take are very similar to those in Fusion Neck, the only difference is in Eq 2. Here, we directly upsample $F_{v 4}$ without element-wise multiplication by global language feature $F_{tg}$ to obtain the $F_{m 4}^{\prime}$. 
 \begin{equation}
F_{m 4}^{\prime}=U p\left(\sigma\left(F_{v 4} W_{v 4}\right)\right),
\end{equation}
The remaining operations are the same as the previous equations. After obtaining the intermediate features $F_{inte }^{\prime}\in \mathbb{R}^{H_{3} \times W_{3} \times C}$ just like $F_{inte}$, we apply three convolution layers to reduce the feature channel dimension size to the number of queries $N_q$.  As a result, we obtain $N_q$ feature maps, which are subsequently flattened in the spatial domain. Specifically, each feature map is flattened to a one-dimensional vector of length $H_{3} \times W_{3}$, where $H_{3}$ and $W_{3}$ are the height and width of the dense vision feature maps, respectively. This results in $F_{vd} \in \mathbf{R}^{N_q\times (H_3W_3)}$ , which contains detailed visual information. And the specific process is formulated as follows:

\begin{equation}
F_{v d}=Flatten\left(Conv\left(F_{inte}^{\prime}\right)\right)^T,
\end{equation}

\textbf{Fused language features}. Our approach incorporates both detailed vision features and holistic global vision features to guide the query generation process. We fuse the language features and the global vision feature by the following equation:
\begin{equation}
F_{t v}=\sigma\left(F_t W_{t}\right)\cdot\sigma\left(F_{vg} W_{vg}\right),
\label{eq:8}
\end{equation}

Here, $F_{t v}\in \mathbb{R}^{L \times C}$, $W_t$ and $W_{vg}$ are two learnable matrices.

\textbf{Multi-Query Generation}. For referring image segmentation, the importance of different words in the same language expression is obviously different. In our approach, we use the dense vision features and the fused language features to generate multiple queries, each corresponding to a different interpretation of the image. We do this by vision-guided attention and computing attention weights between the dense vision features and the fused language features for each query.

In order to derive attention weights for fused language features $F_{tv}$, we incorporate the dense vision features $F_{vd}$. We begin by applying linear projection to $F_{vd}$ and $F_{tv}$. Then, for the $n$-th query ($n = 1,2,...,N_q$), we use the $n$-th dense vision feature vector $f_{vdn} \in \mathbb{R}^{1 \times \left(H_3W_3\right)}$ , along with the fused language features. Specifically, we use $f_{tvi} \in \mathbb{R}^{1 \times C}$ to denote the feature of the $i$-th word ($i = 1,2,...,L$). The attention weight for the $i$-th word with respect to the $n$-th query is computed as the product of projected $f_{vdn}$ and $f_{tvi}$:
\begin{equation}
a_{n i}=\sigma\left(f_{vdn} W_{vd}\right) \sigma\left(f_{tvi} W_{a}\right)^T,
\end{equation}
In the equation, $a_{ni}$ represents a scalar that indicates the importance of the $i$-th word in the $n$-th query, where $W_{vd}$ and $W_{a}$ are learnable matrices. To normalize the attention weights across all words for each query, we apply the Softmax function. The resulting values of $a_{ni}$, after being processed by Softmax, comprise the attention map $A \in \mathbb{R}^{N_q \times L}$. For the $n$-th query, we extract $A_n \in \mathbb{R}_{1 \times L}$ ($n = 1,2,...,N_q$) from A, which represents the emphasis of the words on the $n$-th query. And $A_n$ are used to generate the new queries as the following equation:
\begin{equation}
    F_{qn} = A_n\sigma\left(F_{t v} W_{t v}\right). 
\end{equation} 
The matrix $W_{tv}$ is a learnable parameter and $F_{qn} \in \mathbb{R}^{1 \times C}$ is a new query. The set of all queries comprises the new matrix $F_q \in \mathbb{R}^{N_q \times C}$, which is also vision-language features predominantly influenced by language information. Then the $F_q$ and fused vision features $F_{vt}$ are fed into a standard transformer decoder to obtain fine-grained vision-language features $F_s \in \mathbb{R}^{H_{3} \times W_{3} \times C}$. In the transformer, Q is from $F_{vt}$ and K, V are from $F_q$.

\begin{table*}[thbp]
    \setlength{\belowcaptionskip}{1.0pt}
    \begin{center}
    \caption{Comparisons with the state-of-the-art approaches on three benchmarks.
    We report the results of our method with various backbones. oIoU and mIoU are utilized as the metric.}
    
    \begin{adjustbox}{width=\textwidth}
    \begin{tabular}{l|c|c|c|ccc|ccc|cc}
        \toprule[1.2pt]
        \multirow{2}{*}{Method} & \multirow{2}{*}{Metric} & \multirow{2}{*}{Vision Backbone} & \multirow{2}{*}{Text Backbone} & \multicolumn{3}{c|}{RefCOCO} & \multicolumn{3}{c|}{RefCOCO+} & \multicolumn{2}{c}{G-Ref} \\
        \cline{5-12}
        ~ & ~ & ~ & ~ & val & test A & test B & val & test A & test B & val & test \\
       \midrule
        CGAN~\cite{luo2020cascade}           &\multirow{8}{*}{oIoU} & ResNet-101 & Bi-GRU  &64.86& 68.04& 62.07& 51.03& 55.51& 44.06& 51.01& 51.69\\    
        EFN ~\cite{feng2021encoder}           & & ResNet-101 & Bi-GRU &62.76 & 65.69 & 59.67 & 51.50 & 55.24 & 43.01 & - & - \\
        BUSNet~\cite{9577319}                     & & ResNet-101 & Self-Att& 63.27 &66.41 &61.39& 51.76& 56.87& 44.13&-&-\\
        LAVT~\cite{yang2022lavt}              & & Swin-Base &BERT-Base  &72.73 &  75.82 & 68.79 & 62.14 & 68.38 & 55.10 & 61.24 & 62.09 \\
        
        RefSegformer~\cite{wu2023towards}    & & Swin-Base&BERT-Base& 73.22& 75.64& 70.09 &63.50& 68.69& 55.44& 62.56& 63.07\\
        CrossVLT~\cite{cho2023cross}   & & Swin-Base&BERT-Base&73.44& 76.16& \underline{70.15}& 63.60& 69.10& 55.23& 62.68& 63.75\\
        SADLR~\cite{yang2023semantics}        & & Swin-Base&BERT-Base&\underline{74.24}& \underline{76.25}& 70.06& \underline{64.28} &\underline{69.09}& 55.19& \underline{63.60}& 63.56\\
        \rowcolor{lightgray}EAVL (Ours)                           & & Swin-Base&BERT-Base& \textbf{75.55}&\textbf{77.50}&\textbf{71.53}&\textbf{65.38}&\textbf{70.12}&\textbf{57.75}&\textbf{64.17}&\textbf{64.48}\\
        \midrule
         RPN~\cite{8578700}          & \multirow{11}{*}{mIoU}& ResNet-101  &Bi-LSTM & 55.33 &57.26& 53.93& 39.75& 42.15& 36.11& -& -\\
        RefTr~\cite{li2021referring}           & & ResNet-101  &BERT-Base &70.56 &73.49& 66.57& 61.08& 64.69& 52.73& 58.73& 58.51 \\
         SeqTR~\cite{zhu2022seqtr}           & & ResNet-101  &Bi-GRU &67.26& 69.79& 64.12& 54.14& 58.93& 48.19& 55.67& 55.64 \\
         PKS~\cite{li2023fully}           & & ResNet-101  &Bi-GRU &70.87& 74.23& 65.07& 60.90& 66.21& 51.35& 60.38& 60.98 \\
         CRIS~\cite{wang2022cris}               & & ResNet-101 &Transformer& 70.47 & 73.18 & 66.10 & 62.27 & 68.08 & 53.68 & 59.87 & 60.36 \\

        VLT~\cite{ding2022vlt}                & & Swin-Base&BERT-Base& 72.96& 75.56& 69.60 &63.53& 68.43& 56.92& 63.49& 66.22\\
       
         PVD~\cite{cheng2023parallel}          & & Swin-Base&BERT-Base&74.82& 77.11& 69.52& 63.38 &68.60& 56.92& 63.13& 63.62\\
          
          $\mathbf{M^3}$AII~\cite{liu2023multi}& & Swin-Base&BERT-Base&73.60&76.23&70.36&65.34&70.50&56.98&64.92&\underline{67.37}\\
          FCNet~\cite{yan2024fuse}          & & Swin-Base&BERT-Base&75.02& \underline{77.49}& 71.47& 66.38 &71.25& 58.71& 65.22& 66.49\\
          CRFormer~\cite{Yan_2024}          & & Swin-Base&BERT-Base&\underline{75.26}& 77.38& \underline{71.92}& \underline{66.98} &\underline{71.74}& \underline{59.32}& \underline{65.97}& 66.86\\
        \rowcolor{lightgray}EAVL (Ours)                           & & Swin-Base&BERT-Base& \textbf{77.08}&\textbf{78.52}&\textbf{73.49}&\textbf{67.95}&\textbf{72.17}&\textbf{61.24}&\textbf{67.15}&\textbf{67.68}\\
        \bottomrule[1.2pt]
    \end{tabular}
    \end{adjustbox}
    \label{tab:sota}
    \end{center}
    \vspace{0mm}
\end{table*}

\begin{table}[htbp]
     
    \setlength{\belowcaptionskip}{1.0pt}
    \begin{center}
    
    \caption{Influence of Query Numbers.}
    \setlength{\tabcolsep}{1.8mm}{
    \begin{tabular}{c|c|c|c|c|c|c}
        \toprule[1.2pt]
        $N_q$ & mIoU & Pr@50 & Pr@60 & Pr@70 & Pr@80 & Pr@90 \\
        \midrule
          32 & 63.28 & 73.23 & 69.46 & 63.28 & 49.34 & 16.78 \\
          24 & \textbf{64.85} & \textbf{75.51} & \textbf{71.79} & \textbf{65.22}  & \textbf{51.27}  & \textbf{17.67} \\
          16 & 64.04 & 74.83 & 70.87 & 64.11 & 50.56 & 17.43 \\
          8  & 63.56 & 73.76 & 69.58 & 63.31 & 49.87 & 16.99 \\
          4  & 62.43 & 72.57 & 69.10 & 61.67 & 48.26 & 16.74 \\
          2  & 61.62 & 71.32 & 66.89 & 60.69 & 47.90 & 16.62 \\
          1  & 60.77 & 70.59 & 66.72 & 60.24 & 46.88 & 16.36 \\
        \bottomrule[1.2pt]
    \end{tabular}
    \label{tab:Nq}}
    \end{center}
 
\end{table}

\begin{table}[htbp]
    \begin{center}
    \caption{Comparison of convolutions with a fixed kernel (Fixed) and Vision-Language Aligner (Aligner) in the segmentation stage.}
    \setlength{\tabcolsep}{0.85mm}{
    \begin{tabular}{c|c|c|c|c|c|c|c}
        \toprule[1.2pt]
        $N_q$ & method &mIoU & Pr@50 & Pr@60 & Pr@70 & Pr@80 & Pr@90 \\
        \midrule
          \multirow{2}{*}{24} & Aligner & \textbf{64.85}   &\textbf{75.51}  & \textbf{71.79} & \textbf{65.22} & \textbf{51.27} & \textbf{17.67} \\
            \cline{2-2}
           & Fixed & 63.16  &73.42 & 69.38 & 63.13 & 49.68 & 17.15 \\
          \midrule
          \multirow{2}{*}{16} & Aligner & 64.04 & 74.83 & 70.87 & 64.11 & 50.56 & 17.43 \\
          \cline{2-2}
          ~ & Fixed & 62.58 & 73.04 & 69.45 & 62.09 & 48.58 & 16.84 \\
          \midrule
          \multirow{2}{*}{8} & Aligner & 63.56 & 73.76 & 69.58 & 63.31 & 49.87 & 16.99 \\
          \cline{2-2}
          ~ & Fixed & 61.99 & 72.07 & 68.73 & 61.37 & 48.28 & 16.52 \\
        \bottomrule[1.2pt]
    \end{tabular}
    \label{tab:vl-fix}}
    \end{center}

\end{table}

\begin{table}[thbp]
    \begin{center}
    \caption{Ablation results of global vision feature $F_{vg}$, global language feature $F_{tg}$ and Multi-Query
Estimator on the RefCOCO+ testA set. MQE denotes the Multi-Query Estimator.}
    \setlength{\tabcolsep}{1mm}{
    \begin{tabular}{c|c|c|c|c|c|c|c|c}
        \toprule[1.2pt]
        $F_{tg}$&$F_{vg}$ & MQE & mIoU & Pr@50 & Pr@60 & Pr@70  & Pr@80 & Pr@90 \\
        \midrule
   \checkmark &\checkmark & \checkmark &\textbf{64.85}  &\textbf{75.51} & \textbf{71.79} & \textbf{65.22} & \textbf{51.27}&\textbf{17.67} \\
       \checkmark & \checkmark   & ~ & 63.19  &73.46 &69.43 & 63.21& 49.54  & 17.21\\
         \checkmark &~ & ~ & 62.34  &72.89  &69.27 & 61.86  &48.47 & 16.73 \\
         ~ & ~ &~& 61.76  &71.47  &68.97 & 61.06 &48.39 & 16.60\\
        \bottomrule[1.2pt]
    \end{tabular}
    \label{tab:other_ab}}
    \end{center}
\end{table}

\subsection{Vision-Language Aligner}
In contrast to previous methods~\cite{ding2021vision,yang2022lavt}, which directly use a~\cite{wu2023towards} fixed learned convolution kernel to obtain the result, we leverage the queries and transform them into query-based convolution kernels. Therefore, the kernels are related to the specific language and image input. Specifically, we transform each query into a convolution kernel and use this query-based kernel to do convolution, resulting in a total of $N_q$ masks. This process is conducted by the Multi-Mask Generator, and each mask represents a specific comprehension of the input language expression. Meanwhile, it is desirable to adaptively decide the importance of each mask, allowing the network to focus on the more important and more suitable ones. Specifically, we feed each query into the Multi-Query Estimator, which evaluates its importance and assigns a score reflecting the quality of the mask generated by this query-based kernel. We then use these scores to weigh and sum all the masks, resulting in the final prediction. 

\textbf{Multi-Mask Generator}. As illustrated in Fig.~\ref{fig:method}, Multi-Mask Generator takes $F_s$ and query vectors $F_q$ as input. We extract one query $F_{q n}$ from $F_q$, and $F_{q n}$ is used to generate a query-based kernel. We use a dynamic convolution operation~\cite{chen2020dynamic}, and the parameters of the query-based kernel come from $F_{q n}$. The detailed operations are as follows:
\begin{equation}
\begin{aligned}
& F_p = Up(Conv(Up(F_s))), \\
& F_{p n} = \sigma(W_pF_{q n}),
\end{aligned}
\end{equation}
Here, we use 2 $\times$ upsampling and convolution operation to transform $F_s$ into  $F_p \in \mathbb{R}^{4H_3 \times 4W_3 \times C_p}$, $C_p = \frac{C}{2}$. Then we use a linear layer to transform $F_{q n}$ into $F_{p n} \in \mathbb{R}^{9C_p + 1}$. For the vector $F_{pn}$, we take the first $9C_p$ values as parameters of the $3 \times 3$ convolution kernel whose number of channels is $C_p$, and we take the last value of $F_{pn}$ as bias, resulting in a query-based convolution kernel $W_{q n}$, and then we utilize $W_{q n}$ to do convolution with $F_s$ obtain a mask, which is denoted as $mask_n \in \mathbb{R}^{4H_3 \times 4W_3 \times 1}$. 
\begin{equation}
mask_n = \Psi_n(F_s,W_{q n})
\end{equation}
Here, $\Psi_n$ denotes the $3 \times 3$ convolution with the query-based convolution kernel $W_{q n}$.

\textbf{Multi-Query Estimator}. As illustrated in Fig.~\ref{fig:method}, the Multi-Query Estimator takes the query vectors $F_q$ as input and outputs $N_q$ scores. Each score shows how much the query $F_{q n}$ fits the context of its prediction, and reflects the importance of its response $mask_n$ generated by itself. The Multi-Query Estimator first applies a multi-head self-attention layer and then employs a linear layer to obtain $N_q$ scalars:
\begin{equation}
    S_q = Softmax(W_s(MHSA(F_q))),
\end{equation}
Here, $S_q \in \mathbb{R}^{N_q \times 1}$. The linear layer uses Softmax as an activation function to control the output range. 

The final prediction is derived from the weighted sum of the mask obtained by the Multi-Mask Generator and the score obtained by the Multi-Query Estimator:
\begin{equation}
    y = \sum_{n=1}^{N_q} S_{qn}mask_n.
\end{equation}
Here, $S_{qn}$ is $n$-th scalar of the $S_q$, $y$ denotes the final prediction mask. The model is optimized with cross-entropy loss.
\section{Experiments}
\subsection{Implementation Details}
\textbf{Experiment Settings.} Following previous works~\cite{yang2022lavt}, we utilize Swin-Base~\cite{liu2021swin} and BERT-Base~\cite{lu2019vilbert} as vision and language encoder.  Input images are resized to $480 \times 480$. Input sentences are set with a maximum sentence length of 17 for RefCOCO and RefCOCO+, and 22 for G-Ref. Each Transformer block has 8 heads, and the hidden layer size in all heads is set to 512, and the feed-forward hidden dimension is set to 2048. We conduct training for 40 epochs, utilizing the Adam optimizer with a learning rate of 0.00005 and a polynomial decay~\cite{chen2014semantic}.  We train the model with a batch size of 32 on 8 NVIDIA RTX A6000 with 48 GPU VRAM.

\begin{figure*}[t]
  \centering
  \includegraphics[width=0.95\linewidth]{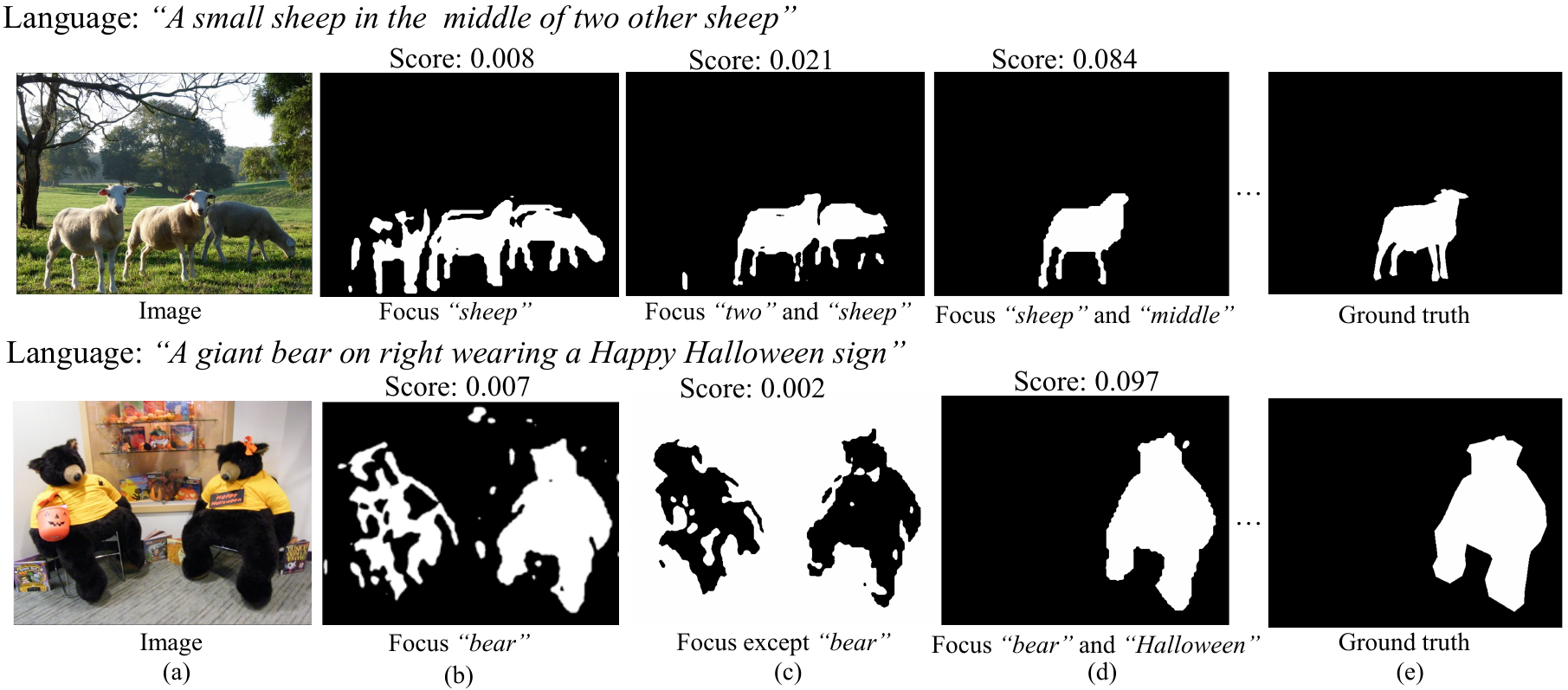}
  \caption{Examples of masks focusing on different regions. Additionally, we provide their cor-
responding scores obtained from the Multi-Query Estimator.}
  \vspace{0pt}
  \label{fig:masks}
\end{figure*}

\begin{table*}[thbp]
    \setlength{\belowcaptionskip}{1.0pt}
    \begin{center}
    \caption{Further comparison with other RIS models. We also apply our Vision-Language Aligner to existing open-source Transformer-based RIS models, demonstrating that our method is a generic plug-and-play module.  }
    
    \begin{adjustbox}{width=0.85\textwidth}
    \begin{tabular}{l|c|ccc|ccc|cc}
        \toprule[1.2pt]
        \multirow{2}{*}{Method} & \multirow{2}{*}{Backbone} & \multicolumn{3}{c|}{RefCOCO} & \multicolumn{3}{c|}{RefCOCO+} & \multicolumn{2}{c}{G-Ref} \\ \cline{3-10}
        
        ~ & ~  & val & test A & test B & val & test A & test B & val & test \\
       \midrule
        CRIS              &\multirow{3}{*}{\shortstack{RN101\\+\\Transformer}} & 70.47 & 73.18 & 66.10 & 62.27 & 68.08 & 53.68 & 59.87 & 60.36 \\
        CRIS + Ours & & 71.25 &73.81 & 66.54& 63.29 & 68.57& 54.67 & 60.78 & 61.22 \\
        EAVL (Ours) & & \textbf{71.98} &\textbf{74.49} & \textbf{67.35}& \textbf{64.07} & \textbf{69.15}& \textbf{55.17} & \textbf{61.21} & \textbf{61.60} \\
        \midrule
         VLT              &\multirow{5}{*}{\shortstack{Swin\\+\\BERT}} &  72.96& 75.96& 69.60& 63.53 &68.43& 56.92& 63.49& 66.22  \\
         LAVT  & & 74.46 & 76.89 & 70.94 & 65.81 & 70.97 & 59.23 & 63.34 & 63.62 \\
        VLT + Ours & & 74.35 &76.87 & 71.74& 64.89 & 69.08& 58.17 & 64.37 & 64.89 \\
        LAVT + Ours & & 75.76 &77.61 & 72.66& 66.77 & 71.65& 60.14 & 64.61 & 65.27 \\
        EAVL (Ours) & & \textbf{77.08}&\textbf{78.52}&\textbf{73.49}&\textbf{67.95}&\textbf{72.17}&\textbf{61.24}&\textbf{67.15}&\textbf{67.68}\\
 
        \bottomrule[1.2pt]
    \end{tabular}
    \end{adjustbox}
    \label{tab:further}
    \end{center}
    \vspace{0mm}
\end{table*}

\textbf{Metrics.} Following previous works ~\cite{ding2022vlt,wang2022cris,yang2022lavt}, we adopt three metrics to verify the effectiveness: overall IoU(oIoU), mean IoU (mIoU) and Precision@$X$. The oIoU is measured as the ratio between the total intersection area and the total union area of all test samples. The mIoU calculates the mean value of per-image IoU over all test samples. oIoU favors larger objects~\cite{yang2022lavt}. The Precision@$X$ measures the percentage of test images with an IoU score higher than the threshold $X \in \{0.5, 0.6, 0.7, 0.8, 0.9\}$, which focuses on the location ability of the method.
\subsection{Datasets}
We conduct our method on three standard benchmark datasets, RefCOCO ~\cite{yu2016modeling}, RefCOCO+ ~\cite{yu2016modeling}, and G-Ref ~\cite{nagaraja2016modeling}, which are widely used in referring image segmentation task. Images in the three datasets are collected from the MS COCO dataset ~\cite{lin2014microsoft} and annotated with natural language expressions. The RefCOCO dataset contains 142,209 referring language expressions describing 50,000 objects in 19,992 images, while the RefCOCO+ dataset contains 141,564 referring language expressions for 49,856 objects in 19,992 images. The main difference between RefCOCO and RefCOCO+ is that RefCOCO+ only contains appearance expressions, and does not include words that indicate location properties (such as left, top, front) in expressions. G-Ref is another prominent referring segmentation dataset that contains 104,560 referring language expressions for 54,822 objects across 26,711 images. The language usage in the G-Ref is more casual and complex, and the sentence length of G-Ref is also longer on average. Furthermore, the G-Ref dataset has two partitions: one created by UMD ~\cite{nagaraja2016modeling} and the other by Google ~\cite{mao2016generation}. In our paper, we report results on the UMD partition.

\subsection{Comparison to State-of-the-Arts}

In Table~\ref{tab:sota}, we evaluate EAVL against state-of-the-art referring image segmentation methods on the RefCOCO, RefCOCO+, and G-Ref datasets. Some of the previous models provide oIoU results, while others provide mIoU, and we use the same metric for comparison. Our proposed method outperforms other methods on all three datasets. When using oIoU as the metric, compared with the second-best RIS model, SADLR~\cite{yang2023semantics}, our method achieves higher performance with absolute margins of \textbf{1.31}, \textbf{1.25}, and \textbf{1.47} points on the validation, testA, and testB subsets of the RefCOCO dataset, respectively. Similarly, our method attains noticeable improvements over the previous state of the art on RefCOCO+ with large margins of \textbf{2.10}, \textbf{1.03}, and \textbf{2.56}. On the G-Ref dataset, our method surpasses the second-best methods on the validation and test subsets from the UMD partition by absolute margins of \textbf{0.57} and \textbf{0.92}. 

Similarly, using mIoU as the metric, compared with the second-best RIS model, CRFormer~\cite{Yan_2024}, our method achieves higher performance with absolute margins of \textbf{1.82}, \textbf{1.14}, and \textbf{1.57} points on the validation, testA, and testB subsets of the RefCOCO dataset, respectively. Additionally, our method attains noticeable improvements over the previous state of the art on RefCOCO+ with large margins of \textbf{0.97}, \textbf{0.43}, and \textbf{1.92}. On the G-Ref dataset, our method surpasses the second-best methods on the validation and test subsets by absolute margins of \textbf{1.18} and \textbf{0.82}.

\subsection{Ablation Study}
We conduct several ablations to evaluate the effectiveness of the key components in EAVL. We do the ablation study on the val split of RefCOCO+, the epoch is set to 25.

\begin{figure*}[t]
  \centering
  \includegraphics[width=0.95\linewidth]{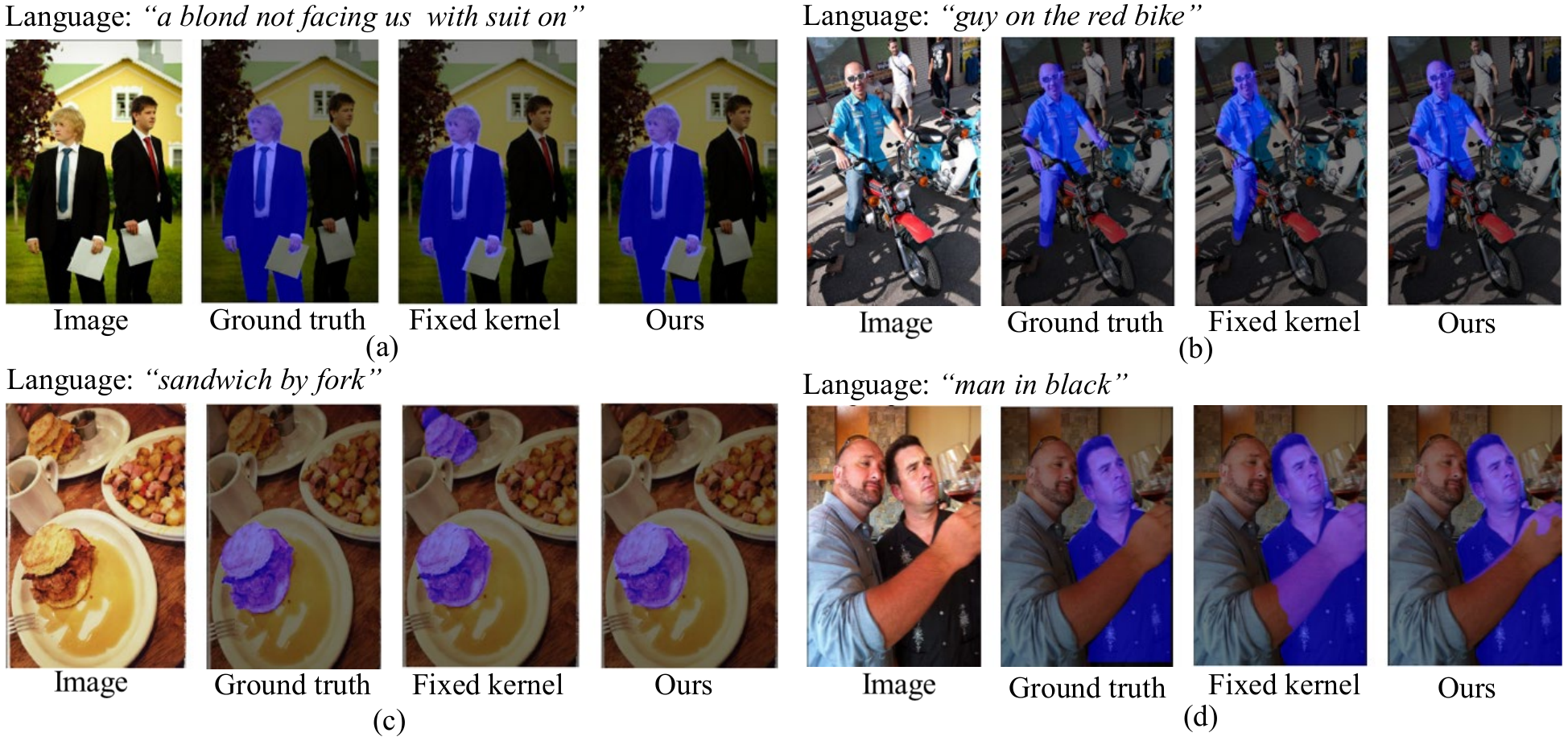}
  \caption{Visualization of our method (using Vision-Language Aligner) and traditional method using convolutions with a fixed kernel in the segmentation stage.}
  \label{fig:results}
\end{figure*}

\begin{table*}[thbp]
    \setlength{\abovecaptionskip}{0pt}
    \setlength{\belowcaptionskip}{0pt}
    \begin{center}
    \vspace{0pt}
    \caption{\textbf{Efficiency comparison with recent models on the val set of RefCOCO+, “FPS” denotes the inference speed.}}
    \adjustbox{scale=1.2}{ 
    \setlength{\tabcolsep}{2.0mm}{
     \begin{tabular}{c|c|c|c|c|c|c|c}
        \toprule[1.2pt]
        Method & Vision Backbone &Text Backbone & mIoU & GFLOPs & time & Params & FPS  \\
        \midrule
          CRIS & ResN101 &Transformer & 62.27&62.72 & 13.7h & 161.25M  & 12.07 \\
          VLT & Swin-Base & BERT-Base&63.53& 130.66  & 59.3h & 204.44M  & 10.71   \\
          LAVT & Swin-Base & BERT-Base&65.81& 124.71  & 54.3h & 203.71M  & 6.69   \\
          \midrule
        Ours & ResN101 & Transformer& 64.07 &99.94& 14.4h & 192.24M  &  11.45\\
        Ours & Swin-Base & BERT-Base & 67.95 &131.01& 58.4h & 202.11M  &  8.78\\
        \bottomrule[1.2pt]
    \end{tabular}}}
    \label{tab:eff}
    \vspace{0pt}
    \end{center}
\end{table*}

\textbf{Query number}. In order to clarify the influence of the query number $N_q$, we set the $N_q$ to various numbers. The results are reported in Table~\ref{tab:Nq}. According to the result, multiple queries can improve the performance of our model which is about $4\%$ from 1 query to 24 queries. However, a larger $N_q$ does not always bring a better result, With the increase of $N_q$, the performance will gradually level off or even decline. Eventually, We select $N_q=24$ due to its best performance on all metrics.


\textbf{Vision-Language Aligner}. The core of our method is utilizing the Vision-Language Aligner. We employ dynamic query-based kernels to replace the fixed kernel. To demonstrate its effectiveness, we conduct a comparison experiment between the Vision-Language Aligner and convolution with a fixed kernel. In this experiment, we replaced the entire Vision-Language Aligner, including the Multi-Mask Generator and Multi-Query Estimator by convolutions with a fixed kernel. In this case, the inputs of the convolutions are only the outputs of the decoder without multiple queries. As shown in Table~\ref{tab:vl-fix}, the results show that under the setting of taking different values of $N_q$, Vision-Language Aligner consistently outperforms conventional fixed convolution kernels evidently. It is noted that both two methods generate multiple queries, but Vision-Language Aligner is more effective thanks that it introduces explicit alignment of language and vision features during the segmentation stage, fully exploring the potential of multiple queries, and thus leading to more accurate text-to-pixel fine-grained correlation. Visualization results are compared in the supplementary material.

\textbf{Multi-Query Estimator.} To demonstrate the effectiveness of our proposed Multi-Query Estimator (MQE), we conduct an experiment by removing it. So the final result is calculated by directly adding all masks with equal weight. As illustrated in Table~\ref{tab:other_ab}, removing the MQE leads to a decrease of 1.02. This indicates that the generated masks need to be aggregated according to the importance of their corresponding queries.

\textbf{Global vision feature and Global language feature.} $F_{vg}$ and $F_{tg}$ are the global representations extracted by the image and text encoder. They are important in the multi-modal fusion stage because they contain the global information of the image and language. By removing the global vision feature $F_{vg}$, we directly use the $F_t$ in Eq.~\ref{eq:8} to generate queries. By removing the global language feature $F_{tg}$, we directly use the $F_{v4}$ in Eq.~\ref{eq:2} to obtain the $F_{m4}$. As shown in Table~\ref{tab:other_ab}, removing either of them will achieve inferior performance than the default setting, and removing both of them get the worst result, demonstrating that global representations are vital for our method and their improvements can be complementary.

\textbf{Further comparison.} To further validate our proposed method, we also used the same backbone as CRIS~\cite{wang2022cris} for the experiment. Additionally, we apply our proposed Vision-Language Aligner to current open-source Transformer-based RIS methods, including CRIS, LAVT~\cite{yang2022lavt}, and VLT~\cite{ding2022vlt}. For example, we implement our approach on CRIS by utilizing features extracted from its encoder to generate queries and applying query-based convolution kernels to replace their fixed kernel on the outputs of the transformer decoder.  As illustrated in Table~\ref{tab:further}, these additional experiments demonstrate that our proposed Vision-Language Aligner is a general approach that can be applied to other transformer-based RIS models.


\section{Visualization}
In this section, we provide several visualization results. We show different masks from the Multi-Mask Generator to demonstrate the mechanism of our approach. We also provide segmentation results to prove the effectiveness of our method.
\subsection{Visualization of multiple masks}
As illustrated in Fig.~\ref{fig:masks}, we present visual examples demonstrating different masks generated by the Multi-Mask Generator, each focusing on distinct regions. These masks are obtained by query-based convolution kernels so that each mask focuses on a different emphasis of the input sentence.  We directly visualize each mask before the weighted sum through a binarization operation. Additionally, we provide their corresponding scores obtained from the Multi-Query Estimator. For instance, in the first picture, when the query emphasizes "sheep," the generated mask depicts three sheep, and the mask score is very low. Only when the query focuses on the keywords "middle" and "shape," the corresponding mask score is the highest. Simultaneously, due to the multiple masks, there are many masks that focus on the correct area, and they also have larger weights. Consequently, after the weighted sum, the masks emphasizing the correct area dominate, resulting in accurate results. That explains why we generate multiple masks and give them corresponding weights.

\subsection{Visualization of segmentation results}
As illustrated in Fig.~\ref{fig:results}, we visualize segmentation results of the ablation studies about the Vision-Language Aligner to demonstrate the effectiveness of this part in our proposed method. We use traditional methods, using convolutions with a fixed kernel as the baseline. We visualize the results of our method and the baseline separately. From the visualization results, we can observe that the absence of the Vision-Language Aligner leads to worse segmentation masks. For instance, as illustrated in Fig.~\ref{fig:results}, the language expression is\emph{"sandwich by fork"}, but the baseline result displays two sandwiches. This is because the baseline method only notices the "sandwich" and does not fully correlate the "by fork" to the image in the pixel level. In our approach, we produce multiple queries, each with distinct emphases, such as queries highlighting \emph{"sandwich"} and \emph{"by fork"}. These queries are transformed into query-based kernels which are then employed for convolution in the segmentation stage, facilitating a comprehensive and explicit alignment between vision and language of varying emphases. As a result, our proposed method establishes a fine-grained correlation between text and pixels, effectively discerning distinctions between a sandwich with a fork and a sandwich without a fork at the pixel level. 

Similarly, as illustrated in Fig.~\ref{fig:results}(a) and Fig.~\ref{fig:results}(b), our method excels in achieving a more explicit alignment between language and image, resulting in better results, particularly in pixel-level details such as the left leg of the blond in Fig.~\ref{fig:results} and the left abdomen of the bike guy in Fig.~\ref{fig:results}(b). These areas were previously disregarded by the baseline method. However, our model is still uncertain in some challenging marginal regions. As illustrated in Fig.~\ref{fig:results}(d), the results of our method also included a small part of another man's hand. We suspect that the reason may be the overlap between the other man's arm and the black man's body, posing challenges for our method in handling such overlaps. This instance shows there is still potential for improvement in our approach, particularly with regard to details and overlapping regions.

\section{Efficiency comparisons}

We conduct efficiency comparisons with current open source methods, using the same training settings, including CRIS\cite{wang2022cris}, LAVT\cite{yang2022lavt} and VLT\cite{ding2022vlt}.
As shown in Table~\ref{tab:eff}, we provide the parameters and GFLOPs of different models. We also report training times using 8 RTX Titan GPUs, along with inference times (image resolution is set to $480 \times 480 $). Compared with previous state-of-the-art methods,  our proposed method achieves better performance while maintaining relatively short training times and fast inference speed.

\section{Conclusion and Future work}
We introduce EAVL, an innovative framework for referring image segmentation (RIS), which leverages a novel Vision-Language Aligner to supplant convolutions with a fixed kernel in the segmentation stage. By generating a series of queries and transforming them into query-based convolution kernels, the convolution kernels are directly related to the input language and image. Our approach explicitly aligns the vision and language, resulting in an effective text-to-pixel fine-grained correlation. Experimental results demonstrate that EAVL significantly outperforms previous state-of-the-art methods across the RefCOCO, RefCOCO+, and G-Ref datasets.  Moreover, our proposed Vision-Language Aligner is a flexible and generic plug-and-play module that can be seamlessly integrated into existing transformer-based RIS models, leading to substantial performance gains.

Our approach provides a novel solution for explicitly aligning vision and language features in RIS, paving the way for new research directions in this field. Cross-modal alignment technology is essential in many other multi-modal tasks, including Referring Expression Comprehension~\cite{subramanian2022reclipstrongzeroshotbaseline,yu2018mattnet,chen2018real}, Referring Video Object Segmentation~\cite{10495702,wu2022languagequeriesreferringvideo,tang2023temporalcollectiondistributionreferring}, and AI-generated content (AIGC)~\cite{ramesh2021zero,li2018video,khachatryan2023text2video}. Our future research should aim to extend this methods, exploring explicit cross-modal alignment techniques in these tasks. Such advancements could significantly improve accuracy and generalization in cross-modal systems, making valuable contributions to the broader multi-modal AI community.

\vfill
\bibliographystyle{IEEEtran}
\bibliography{Reference}

\end{document}